\documentclass[10pt,twocolumn,letterpaper]{article}

\usepackage{iccv}              
\usepackage{kotex}

\usepackage[dvipsnames]{xcolor}

\usepackage{multirow} 
\usepackage{amsmath} 
\usepackage{bbm} 
\usepackage{siunitx} 
\usepackage{kotex} 
\usepackage{array,multirow,graphicx}
\usepackage{float}
\usepackage{amsmath}
\usepackage{mathtools}
\usepackage{bbm}
\usepackage{hhline}
\usepackage{boldline}
\usepackage{tabularx}
\usepackage{float}
\usepackage{comment}
\usepackage{color}
\usepackage{amssymb}
\usepackage{booktabs}
\usepackage{multicol}
\usepackage{microtype}      
\usepackage{xcolor}         
\usepackage{algorithm}
\usepackage{algpseudocode}
\usepackage{pifont}
\usepackage{soul} 
\usepackage{subcaption}
\usepackage{wrapfig}
\usepackage{lipsum}
\usepackage{booktabs}
\usepackage{xcolor,colortbl}
\usepackage{nicematrix}
\usepackage{kotex}
\usepackage{dsfont}
\usepackage{caption}
\usepackage[accsupp]{axessibility}
\newcommand\blfootnote[1]{%
  \begingroup
  \renewcommand\thefootnote{}\footnote{#1}%
  \addtocounter{footnote}{-1}%
  \endgroup
}

\definecolor{iccvblue}{rgb}{0.21,0.49,0.74}
\usepackage[pagebackref,breaklinks,colorlinks,allcolors=iccvblue]{hyperref}

\def\paperID{4155} 
\def\confName{ICCV}
\def\confYear{2025}

\title{Selective Contrastive Learning for Weakly Supervised Affordance Grounding}

\author{
    WonJun Moon$^{\dagger}$ \;\;\; Hyun Seok Seong$^{\dagger}$ \;\;\; Jae-Pil Heo$^{\ast}$ \\
    Sungkyunkwan University \\
{\tt\small \{wjun0830, gustjrdl95, jaepilheo\}@skku.edu}
}

\begin{document}
\maketitle
\begin{abstract}
Facilitating an entity's interaction with objects requires accurately identifying parts that afford specific actions.
Weakly supervised affordance grounding~(WSAG) seeks to imitate human learning from third-person demonstrations, where humans intuitively grasp functional parts without needing pixel-level annotations.
To achieve this, grounding is typically learned using a shared classifier across images from different perspectives, along with distillation strategies incorporating part discovery process.
However, since affordance-relevant parts are not always easily distinguishable, models primarily rely on classification, often focusing on common class-specific patterns that are unrelated to affordance.
To address this limitation, we move beyond isolated part-level learning by introducing selective prototypical and pixel contrastive objectives that adaptively learn affordance-relevant cues at both the part and object levels, depending on the granularity of the available information.
Initially, we find the action-associated objects in both egocentric~(object-focused) and exocentric~(third-person example) images by leveraging CLIP.
Then, by cross-referencing the discovered objects of complementary views, we excavate the precise part-level affordance clues in each perspective.
By consistently learning to distinguish affordance-relevant regions from affordance-irrelevant background context, our approach effectively shifts activation from irrelevant areas toward meaningful affordance cues.
Experimental results demonstrate the effectiveness of our method.
Codes are available at \href{github.com/hynnsk/SelectiveCL}{github.com/hynnsk/SelectiveCL}.

\blfootnote{
$^\dagger$ Equal contribution} 
\blfootnote{
$^\ast$ Corresponding author
}
\end{abstract}    
\section{Introduction}
\label{sec:intro}
Humans learn to interact with objects by observing others and recognizing relevant object parts in interactions~\cite{li2023locate, ardon2019learning}. 
Similarly, weakly supervised affordance grounding focuses on identifying which parts of an object afford particular interactions within the environment in which humans typically learn~\cite{AGD20K, nagarajan2019grounded, li2023locate, wsma, qian2024affordancellm, jang2025intra, qian2023understanding}.
Specifically, a target egocentric image~(object-focused) is provided with an action class name and a few exocentric images~(human-object interaction examples given in third-person view) to localize the affordable parts within egocentric image~\cite{AGD20K, nagarajan2019grounded, qian2024affordancellm}.
Then, the model is trained to localize action-affordable parts when an egocentric image is provided along with an action class.

\begin{figure}
    \centering
    \vspace{-0.2cm}
    {\includegraphics[width=\linewidth]{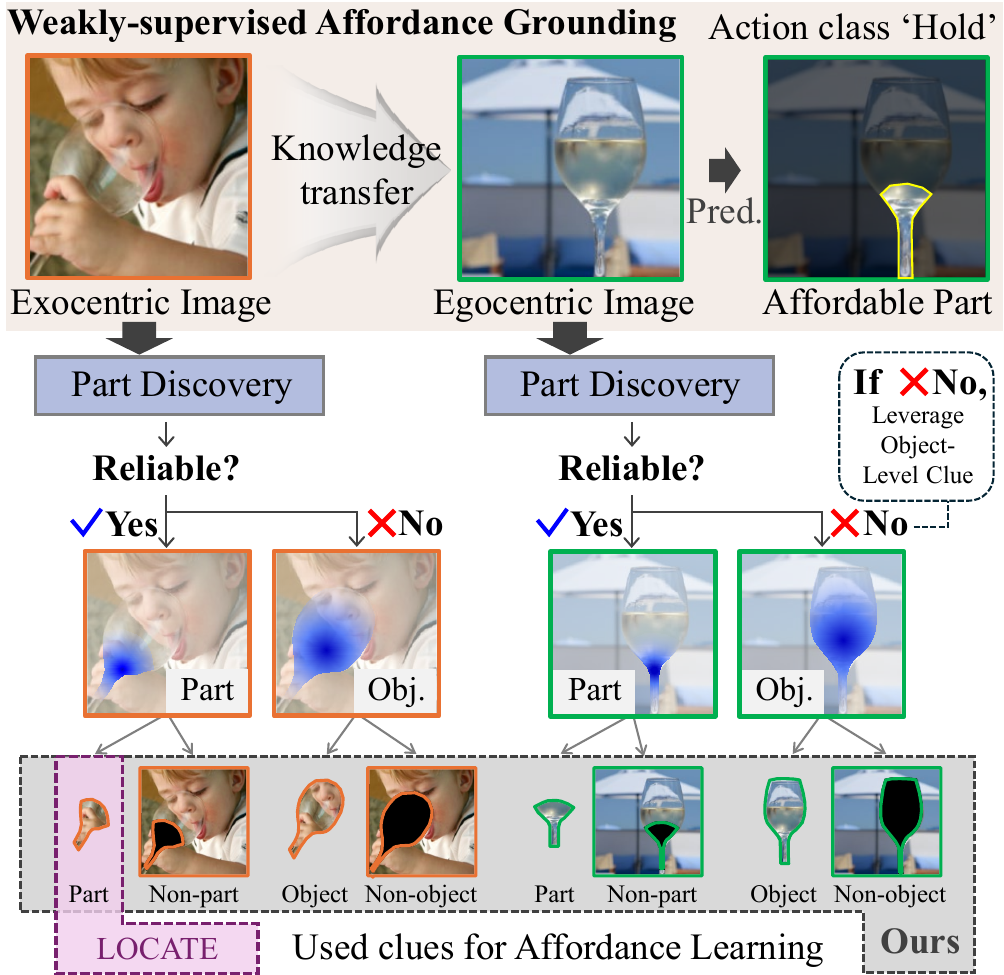} }
    \vspace{-0.45cm}
    \caption{
        (Up) Goal of WSAG is to identify action-affordable parts within the egocentric image, given exocentric images as contextual hints. 
        (Down) To perform affordance learning, we first discover the part-relevant clues from both egocentric and exocentric images. 
        When these parts are deemed reliable in representing affordance-relevant regions, the model learns to distinguish these parts from the other parts. 
        If not, we instead utilize object-level clues to distinguish objects from the background.
        Compared to our baseline~(LOCATE~\cite{li2023locate}), which only exploits reliable parts of exocentric images, our approach extends affordance learning to learn from both egocentric and exocentric views and also both from affordance-relevant and affordance-irrelevant clues.
        By leveraging all these types of clues within the mini-batch at once, the model learns to distinguish affordance-relevant parts from representations of other affordance classes and backgrounds.
        %
    }
\label{fig_approach}
\vspace{-0.3cm}
\end{figure}


In this vein, knowledge distillation is widely studied~\cite{li2023locate,wsma}, along with action classification to produce a class activation map~(CAM)~\cite{cam} for localization.
For example, LOCATE~\cite{li2023locate} introduced a part-level distillation approach. 
It extracts action-affordable parts from exocentric images by segmenting interaction-involved regions identified by CAM. 
These action-affordable parts are distilled into the egocentric image representations only when they are precisely identified, enabling alignment with affordance-relevant regions.

Yet, as training without dense annotation progresses, they tend to locate distinguishable parts necessary for action classification even if they are not directly related to the affordable part.
%
This is because affordance-relevant clues are not always clearly distinguishable, thus, the distillation is only applied intermittently.
To address this, we go beyond solely focusing on part feature distillation; our primary objective is to consistently provide contextual cues to distinguish between affordance-relevant and affordance-irrelevant representations.
The overall intuition is illustrated in Fig.~\ref{fig_approach}.

We begin by collecting object-level affordance-relevant clues from both egocentric and exocentric images, then gradually refine them to part-level clues.
The model is then trained to focus on these affordable parts via dedicated selective contrastive learning.
Specifically, if the identified part clue is deemed to correspond to an affordance region reliably, the model learns to distinguish it from other irrelevant parts. 
Conversely, if the identified part is deemed unsuitable, the model is trained to distinguish the target object clue~(identified using the object affinity map) from the background, preventing attention to affordance-irrelevant regions.
First, to collect the clues for action-associated objects, we leverage CLIP~\cite{CLIP} to generate an object affinity map that encompasses affordance-relevant parts. 
The identified target object then serves as a basis for discovering part-level affordance clues.
For part discovery within the exocentric view, we refine the part discovery algorithm from LOCATE~\cite{li2023locate} by leveraging the target object to improve precision. 
Specifically, the object affinity map is used to filter out object-irrelevant part candidates, ensuring that the affordance-relevant parts belong to the target object.
Conversely, to extract part clues in the egocentric view, we exploit the properties of foundation models that CLIP tends to be more responsive to prominent objects~\cite{CLIPReward}.
Specifically, we assess part cues by analyzing the difference in model activation between egocentric and exocentric images, where responses tend to be weaker in exocentric images due to smaller object scales and occlusions.

Upon gathering object and part clues from both views, we design two types of contrastive learning to leverage the collected affordance-relevant clues.
Initially, we propose prototypical contrastive learning to exploit affordance-relevant clues from the exocentric view, offering several key advantages over previously used pairwise distillation strategies~\cite{li2023locate, wsma}.
While pairwise distillation focuses solely on reducing the distance between representations of paired egocentric and exocentric images, prototypical contrastive learning not only encourages egocentric-exocentric aligned representations but also distinguishes each prototype from diverse background information and the prototypes of other action classes. 
This enables the model to capture more discriminative representations specific to each action class.
On the other hand, pixel-level contrastive learning further optimizes the localization of affordable parts with precise pixel-level clues.
Specifically, it directly uses affordance-relevant clues in egocentric images to disentangle affordance-relevant pixels from the others in each image.
This facilitates pixel representations to be distinguished based on their affordance relevance at the level of gathered clues.

To sum up, our contributions are:
(i) We propose prototypical contrastive learning to benefit part representation learning by leveraging the semantics of other action classes and backgrounds.
(ii) We propose pixel contrastive learning to supplement the fine-grained localization of affordance-relevant regions.
(iii) We present a post-processing step to calibrate CAM prediction by leveraging CLIP's capability to detect text-specified objects.
Our approach consistently outperforms 
(iv) Our approach demonstrates superior performance over prior methods, particularly in challenging unseen scenarios that closely reflect real-world conditions.
\section{Related Work}

\label{sec:relatedwork}
\subsection{Visual Affordance Grounding} 
Visual affordance grounding aims to locate the responsible object parts to certain actions~\cite{li2024one}.
To minimize the gap between perception and action, extensive attention is being put into affordance grounding among the researchers of computer vision and robotics~\cite{li2023locate, hassanin2021visual, kokic2017affordance}.
Initially studied in a supervised setting~\cite{myers2015affordance, nguyen2017object, chuang2018learning}, affordance grounding is recently being studied more in weakly supervised scenarios where costly dense annotations are not required~\cite{wsma, qian2024affordancellm, jang2025intra, chen2024worldafford, AGD20K}.
For example, LOCATE~\cite{li2023locate} uses CAM to identify interaction-involved regions and applies K-means clustering to find the affordance-relevant parts in exocentric images for distillation.
WSMA~\cite{wsma} exploited the semantics of CLIP through an attention mechanism to address the limitation of discrete classification labels in illustrating the semantics of actions.
Also, more recent works utilize diverse foundation models, such as ALBEF~\cite{li2021align}, SAM~\cite{kirillov2023segment}, LLAVA~\cite{liu2024visual}, and GPT~\cite{achiam2023gpt}, to obtain part-level knowledge~\cite{chen2024worldafford, qian2024affordancellm, jang2025intra, rai2024strategies}.
Our approach, despite not relying on recent foundation models, significantly outperforms them by effectively addressing cases where reliable parts cannot be identified and leveraging background context to prevent the model from focusing on affordance-irrelevant regions.



\subsection{Weakly Supervised Object Localization}
Weakly Supervised Object Localization~(WSOL) aims to localize objects using only image-level labels.
Conventionally, CAM-based methods have been widely studied due to their effectiveness~\cite{zhang2018adversarial, wei2022weakly, xue2019danet, mai2020erasing, choe2019attention, yasuki2024cam, zhao2023generative}.
Yet, these often suffer from shortcut learning~\cite{geirhos2020shortcut}, which limits CAM coverage, making CAM expansion a common strategy for WSOL.
For example, HaS~\cite{kumar2017hide} randomly masks image patches during training, CutMix~\cite{yun2019cutmix} enhances masked images, and LoRot~\cite{moon2022tailoring} introduces pretext tasks involving random scaling and positioning to broaden the model's receptive field.
A similar challenge arises in WSAG, where the goal is to localize affordance-relevant parts for a given action class, independent of object categories.
Although WSAG also struggles with the model focusing on commonly appearing details within each action class, CAM expansion is not always a suitable solution, as affordance-relevant parts are often small.
To address this, we propose a selective strategy that adaptively determines whether to expand the CAM to the object region when a reliable part cannot be identified or to concentrate CAM activation when a reliable part is available.


\subsection{Contrastive Learning}
Contrastive learning pulls together instances with positive relationships while pushing apart those with negative relationships~\cite{simclr, moco, swav, pirl}. 
It has been employed in various fields by adapting the criteria for determining the relationships between instances.
For instance, augmented pairs of the same instance are regarded as positives in an unsupervised setting~\cite{simclr}, while samples within the same class are treated as positives in a supervised setting~\cite{supclr}.
For  WSAG, LLM has been employed to derive the relationships between interaction types~\cite{jang2025intra}.
Also, there is significant variation in the units to which contrastive learning is applied.
For example, while the images are typical units~\cite{simclr, simsiam}, prototypes~\cite{li2020prototypical}, pixels~\cite{flsl, seong2023leveraging, seong2024progressive, sung2024contextrast, zhao2021contrastive}, or even the similarity between modalities~\cite{CLIP} are popular sources.
In this work, we introduce prototypical and pixel contrastive learning that adaptively selects the training level to optimize both object- and part-level regions.



\section{Method}
\label{sec:method}
\begin{figure}[t]
\vspace{-0.cm}
    \centering
    {\includegraphics[width=1.\columnwidth]{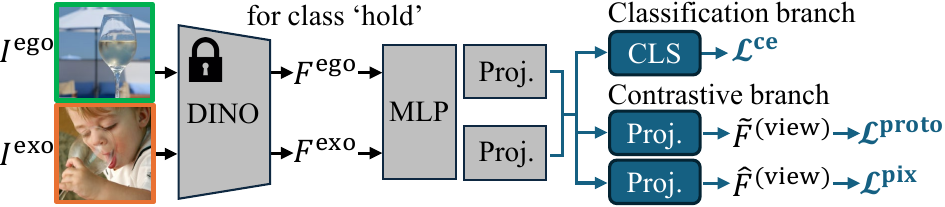} }
    \vspace{-0.2cm}
    \caption{
        Overall flow. Egocentric and exocentric images are processed to perform classification and selective contrastive learning.
        Note that (view)$\in$\{ego, exo\}.
    }
\label{fig_overall_flow}
\vspace{-0.cm}
\end{figure}
\subsection{Method Overview}
In Fig.~\ref{fig_overall_flow}, we illustrate the overall framework.
Given a pair of an egocentric image ${I}^{\text{ego}}$ and multiple exocentric images ${I}^{\text{exo}}$, the inputs are processed using DINO~\cite{DINO} followed by projection layers.
Then, for prototypical and pixel contrastive learning, features are further projected to obtain ${\tilde{F}}^{\text{(view)}}$ and ${\hat{F}}^{\text{(view)}}$, respectively, where (view) $\in$ \{ego, exo\}.
We note that the features of the egocentric image are represented as ${\tilde{F}}^{\text{ego}}, {\hat{F}}^{\text{ego}} \in \mathbb{R}^{B \times H \times W \times D}$, while the features of the exocentric images are given by ${\tilde{F}}^{\text{exo}}, {\hat{F}}^{\text{exo}} \in \mathbb{R}^{B \times E \times H \times W \times D}$, where $B$ denotes the batch size, $H$ and $W$ represent the spatial dimensions, $D$ is the feature dimension, and $E$ indicates the number of exocentric images.
While the contrastive learning branch focuses on learning affordance knowledge within the egocentric images, the classification branch with the shared classifier captures the shared semantic information between egocentric and exocentric views.
For inference, CAM $C^{\text{ego}}$ is derived from classification branch using only egocentric images and affordance text prompts.

To conduct selective contrastive learning, we first establish target supervision by identifying action-associated objects in Sec.~\ref{object_discovery}.
Subsequently, in Sec.~\ref{protoCL} and Sec.~\ref{pixelCL}, we introduce prototypical and pixel contrastive learning, respectively, along with the part-level target discovery process.



\begin{figure}[t]
    \centering
    \vspace{-0.1cm}
    {\includegraphics[width=1.\columnwidth]{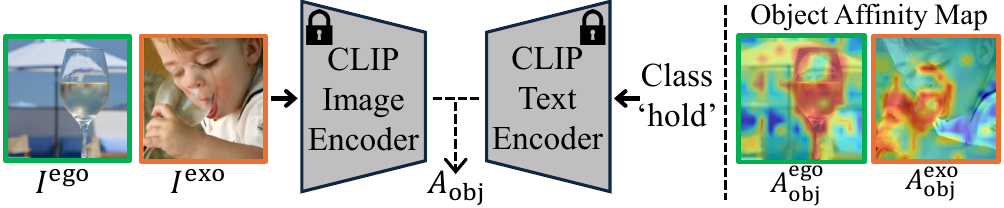} }
    \vspace{-0.4cm}
    \caption{
        Illustration of object discovery. 
        The object affinity map is derived from CLIP as a zero-shot image-text similarity map.
    }
\label{fig_object_discovery}
\vspace{-0.3cm}
\end{figure}

\subsection{Object Discovery}
\label{object_discovery}
As illustrated in Fig.~\ref{fig_object_discovery}, we leverage CLIP to define the object affinity map.
Particularly, we employ the strategy of ClearCLIP~\cite{lan2024clearclip} to enhance local discriminability in visual features.
Given egocentric features and exocentric features from CLIP visual encoder, we calculate cosine similarity with CLIP text features of action prompt to obtain an object affinity map for each perspective, namely $A^{\text{ego}}_\text{obj}\in\mathbb{R}^{B \times H\times W}$ and $A^{\text{exo}}_\text{obj}\in\mathbb{R}^{B\times E\times H\times W}$~(Details for action prompts are in Appendix).
Note that the term \textit{object affinity map} is derived from its characteristic to highlight affordance-relevant objects when action prompt is given, as shown in Fig.~\ref{fig_vis_pixelpart}.

\begin{figure}[t]
    \centering    {\includegraphics[width=1.\columnwidth]{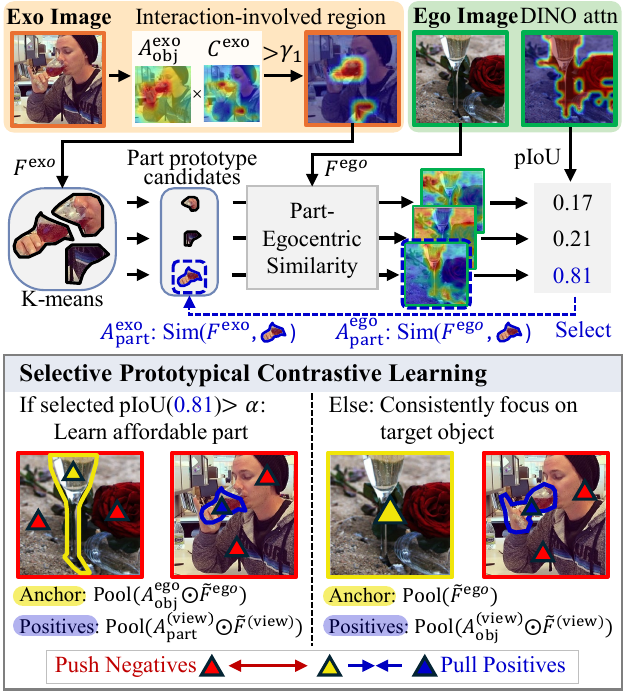} }
    \vspace{-0.55cm}
    \caption{
        Illustration of prototypical contrastive learning.
        (Up) Process of identifying part clues in exocentric images.
        Discovered objects are segmented to extract part candidates, which are then matched with DINO's attention map.
        (Down) Prototypical contrastive learning is selectively applied based on the reliability of part clues. 
        When reliable, object anchors in egocentric images are attracted toward part clues, but otherwise image anchors are drawn toward object clues in exocentric images.
    }
\label{fig_prototypical_contrastive_learning}
\vspace{-0.45cm}
\end{figure}

\subsection{Prototypical Contrastive Learning}
\label{protoCL}
Prototypical contrastive learning operates upon gathered affordance-relevant clues within exocentric view.
Simply put, prototypes for affordable parts in exocentric images are distilled towards corresponding prototypes within egocentric images via contrastive learning.
The key advantage of prototypical contrastive approach over previous works is that it encompasses the process of learning negative relationships between prototypes.
Thus, we claim that the classification bias towards affordance-irrelevant regions, \textit{i.e.}, background context, and non-affordable object parts, can be mitigated.

\noindent\textbf{Part-level Clues within Exocentric Images.}
We begin by illustrating the process of gathering part-level clues in exocentric images, as shown in Fig.~\ref{fig_prototypical_contrastive_learning}.
Specifically, we adapt the algorithm used from previous work~\cite{li2023locate}.
To illustrate, whereas previous work directly thresholded CAM prediction $C^\text{exo}$ to identify interaction-involved regions in exocentric images, we first combined $C^\text{exo}$ with the object affinity map $A^{\text{exo}}_{\text{obj}}$ before applying a threshold $\gamma_1$.
This ensures that the region of interest is constrained to object regions, mitigating the risk of imprecise CAM predictions and improving the affordance relevance of the extracted affordance cues.
The rest of the process follows that of previous work~\cite{li2023locate}.
First, based on the intuition that CAM regions consist of background, affordance-relevant part, and other elements, K-means clustering~(K=3) is applied.
Then, the centroids~(candidates of part prototype) are compared with the egocentric DINO~\cite{DINO} feature $F^{\text{ego}}$ to generate part-egocentric similarity maps.
These maps are then assessed to determine whether each corresponding centroid represents the affordance-relevant parts by comparing them with the self-attention map of the egocentric image from DINO~\cite{DINO}, measured by pIoU~\cite{li2023locate}~(DINO attention map can be replaced by object affinity map).
Finally, only the centroid corresponding to the highest pIoU that exceeds a threshold $\alpha$ is selected as the designated part.
If either condition is not met, the part~(centroid) is considered unreliable and excluded from training.
Consequently, for instances with reliable part prototype, part affinity maps $A^{(\text{view})}_{\text{part}}$ are defined as the similarity between the selected part prototype and spatial features~(i.e., ${F}^{\text{ego}}$ and ${F}^{\text{exo}}$).

\noindent\textbf{Selective Prototypical Contrastive Learning.}
Due to the inconsistent availability of affordable part clues, the typical approach is to exploit the knowledge in exocentric images only when the reliable part is discovered.
Yet, this triggers the affordance grounding task to become heavily reliant on classification tasks, which is vulnerable in capturing target object parts since its goal is to find the most discriminative features for action classification.

Therefore, we design a loss function that consistently leverages the knowledge of interaction-involved regions in exocentric images throughout the training.
Specifically, our prototypical contrastive learning integrates learning-level selectivity for both the target and anchor representations.
When the discovered part prototype within exocentric images is deemed reliable, we use it as the target prototype for distillation into the egocentric object prototype. 
Otherwise, we define the object prototype to serve as the target and the entire egocentric image as an anchor.
This design, which sets the object prototype as the default distillation target, encourages the model to consistently focus on the target object while disregarding background context in egocentric images. 
Furthermore, when part supervision is available, it reinforces attention to affordable parts, enhancing the model’s ability to capture details of affordable parts.


To leverage the object/part clues in prototypical contrastive learning, we initially construct prototypes.
Particularly, four types of prototypes, namely $P^{\text{ego}+}, P^{\text{ego}-}, P^{\text{exo}+}$ and $P^{\text{exo}-}$, are produced which refer to the positive and negative prototypes of object/part clues in each view depending on the level of gathered clues.
In particular, these positive and negative prototypes are constructed with following functions~($\Phi^+$ and $\Phi^-$) using instance feature $Z \in \mathbb{R}^{H\times W\times D}$, target clue $M \in \mathbb{R}^{H\times W}$ and CAM prediction $C \in \mathbb{R}^{H\times W}$:
\begin{equation}
\begin{split}
\label{Eq_prototype_process}
    & \!\!\Phi^{+}(Z,M)= \text{norm}(\text{Pool}( Z \odot M ) ),
    \\
    & \!\!\Phi^{-}(Z,M,C)= \text{norm}(\text{Pool}( Z \odot (\beta -M\odot C) ) ),
\end{split}
\end{equation}
where norm($\cdot$) indicates Frobenius normalization along channel axis, Pool($\cdot$) denotes spatial average pooling, and $\beta$ is a bias term to prevent training instability incurred by imprecise CAM $C$ at the initial training epoch.
Note that $\odot$ is defined as ($\mathbf{X} \odot \mathbf{Y})_{\text{i},\text{j},\text{k}} = \left( x_{\text{i},\text{j},\text{k}} \right)\times \left( y_{\text{i},\text{j}} \right), \quad \forall \text{i} \in \{1, \dots, H\}, \forall \text{j} \in \{1, \dots, W\}, \forall \text{k} \in \{1, \dots, D\}$ to apply Hadamard product between $\mathbf{X}$ and $\mathbf{Y}$ in different shapes. 
In short, the positive prototype maintain a consistent focus on target regions by masking with target clue $M$, which is often more precise than CAM prediction, while the background prototype captures general background semantics and unaffordable parts.

Subsequently, let $\mathds{I}$ denote the index set of both the exocentric and egocentric instances within the mini-batch which represents instances with precise part-level prototypes~(we assume that there is only one exocentric image per egocentric image in this subsection, thereby $\mathds{I}$ can be shared for simplicity).
Then, the egocentric anchor $z^{\text{ego}}_{b}$ of $b$-th instance and the prototypes are formed as follows:
\begin{equation}
    \begin{split}
    \!\!\!z^{\text{ego}}_{b} \!&= 
    \begin{cases} 
      \Phi^+(\tilde{F}^\text{ego}_{b}\!,A^\text{ego}_\text{obj, b}) & \text{if } b \in \mathds{I}, \\
      \text{norm}(\text{Pool}(\tilde{F}^{\text{ego}}_b)) & \text{otherwise},
    \end{cases} \\
    \!\!\!P^{(\text{view})+}_{b} \!&= 
    \begin{cases} 
      \Phi^{+}(\tilde{F}^{(\text{view})}_{b}, A^{(\text{view})}_{\text{part}, b}) & \text{if } b \in \mathds{I}, \\
      \Phi^{+}(\tilde{F}^{(\text{view})}_{b}, A^{(\text{view})}_{\text{obj}, b}) & \text{otherwise},
    \end{cases} \\
    \!\!\!P^{(\text{view})-}_{b} \!&= 
    \begin{cases} 
      \Phi^{-}(\tilde{F}^{(\text{view})}_{b}, A^{(\text{view})}_{\text{part}, b}, C^{(\text{view})}) & \text{if } b \in \mathds{I}, \\
      \Phi^{-}(\tilde{F}^{(\text{view})}_{b}, A^{(\text{view})}_{\text{obj}, b}, C^{(\text{view})}) & \text{otherwise}.
    \end{cases}
    \end{split}
\end{equation}

Consequently, our selective prototypical contrastive learning for $b$-th instance in mini-batch is expressed as:
\begin{equation}
    \mathcal{L}^{\text{proto}}_{b} = \frac{-1}{|\mathbf{P}^{+}_{b}|}\sum_{p \in \mathbf{P}^{+}_{b}} \log \frac{\exp(z^{\text{ego}}_{b} \circ p / \tau)}{\!\!\sum\limits_{n \in (\mathbf{P}^{+}_{b} \cup \mathbf{P}^{-}_{b}) )} \!\! \exp(z \circ n / \tau)},
\end{equation}
where $\circ$ and $\tau$ denote dot product and temperature parameter, respectively. $\mathbf{P}^{+}_{b}$ and $\mathbf{P}^{-}_{b}$ which represent the sets of positive and negative prototypes for $b$-th instance are defined as:
\begin{equation}
\label{proto_pos_set}
    \begin{split}
    \!\!\mathbf{P}^{+}_{b} \! =\! \bigcup_{(\text{view})} \bigcup_{i\in\mathcal{B}}
    \{ P^{(\text{view})+}_{i} | \delta(P^{(\text{view})+}_{i})  = \delta(z^{\text{ego}}_{b}) \},
    \end{split}
\end{equation}
\begin{equation}
\label{proto_neg_set}
    \begin{split}
    \!\!\!\mathbf{P}^-_{b} \! = \!\!\!\bigcup_{(\text{view})}& \biggl\{ \bigcup_{i\in\mathcal{B}}
    \{ P^{(\text{view})-}_{i} | \delta(P^{(\text{view})-}_{i})\!  = \delta(z^{\text{ego}}_{b}) \} \\ 
    &\bigcup_{j\in\mathcal{B}}
    \{ P^{(\text{view})+}_{j} | \delta(P^{(\text{view})+}_{j}) \ne\delta(z^{\text{ego}}_{b}) \}\! \biggl\},\!
    \end{split}
\end{equation}
where $i$, $j$ denote the index from batch index set $\mathcal{B}$, and $\delta$ is a function to output the action class label of given instances.
Consequently, prototypical contrastive learning directs the model’s activation toward affordance-relevant regions. 
Specifically, object-level learning enhances focus on object regions, and when affordance-relevant parts are present, it further refines features to capture part-specific information within the object.

\subsection{Pixel Contrastive Learning}
\label{pixelCL}
In prototypical contrastive learning, we encourage the model to prioritize foreground objects over the entire image and, within these objects, to focus on specific parts. 
However, we remark that only the implicit guidance is provided on each pixel of affordable parts.
Thus, we additionally propose pixel contrastive learning to supplement the fine-grained localization capability by learning correspondences between pixels in each egocentric image.

\noindent\textbf{Part-level Clues within Egocentric Images.}
Symmetrically to the use of egocentric view for gathering part clues in exocentric images, we utilize exocentric view as contextual cues to capture part clues in egocentric images.
Specifically, we leverage the property of foundation models~(CLIP) that they are more responsive to salient objects~\cite{CLIPReward}.
Thus, we expect stronger activations for affordable parts in egocentric images compared to their exocentric counterparts when matched with text prompts describing the part to perform a specific action with.
This is because exocentric images depict objects in use, often capturing them at a small scale and making them more susceptible to occlusions.

\begin{figure}
    \centering
    \vspace{-0.2cm}
    {\includegraphics[width=0.97\linewidth]{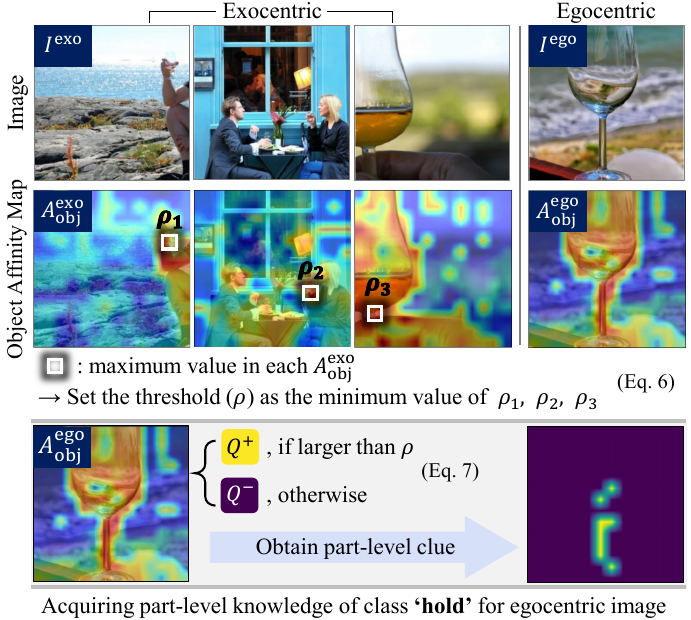} }
    \vspace{-0.3cm}
    \caption{
        An illustration of binarizing objects within egocentric images based on affordance criterion. The most salient pixel in each exocentric object affinity map serves as a reference, establishing a criterion to classify each pixel in the egocentric image as part of an affordable region~($Q^+$) or a non-affordable region~($Q^-$).
        The minimum value among $\rho_1, \rho_2$, and $\rho_3$ is used as criterion.
    }
\label{fig_pixel}
\vspace{-0.3cm}
\end{figure}

\begingroup
\setlength{\tabcolsep}{5.5pt} 
\renewcommand{\arraystretch}{0.90} 
\begin{table*}[t]
\centering
\small
{ 
\vspace{-0.25cm}
 \caption{Performance comparison on the AGD20K and HICO-IIF datasets.}
\label{table_main}
\vspace{-0.3cm}
 \begin{tabular}{l|c|ccc|ccc|ccc}
 \hlineB{2.5}
 \multicolumn{1}{c|}{\multirow{2}{*}{Method}} & \multicolumn{1}{c|}{\multirow{2}{*}{Model}} & \multicolumn{3}{c|}{AGD20K-Seen} & \multicolumn{3}{c|}{AGD20K-Unseen} & \multicolumn{3}{c}{HICO-IIF} \\
\multicolumn{1}{l|}{} & \multicolumn{1}{l|}{}& KLD$\downarrow$ & SIM$\uparrow$ & NSS$\uparrow$ & KLD$\downarrow$ & SIM$\uparrow$ & NSS$\uparrow$ & KLD$\downarrow$ & SIM$\uparrow$ & NSS$\uparrow$ \\ \hlineB{2.5}
\rowcolor{gray!40}
\multicolumn{11}{c}{Zero-Shot Vision-Language Model} \\
Clear-CLIP~\cite{lan2024clearclip} & CLIP & 1.573 & 0.294 & 0.945 & 1.723 & 0.262 & 0.976 & 1.746 & 0.252 & 1.032 \\ 
\rowcolor{gray!40}
\multicolumn{11}{c}{Weakly Supervised Object Localization} \\
SPA~\cite{pan2021unveiling} & - & 5.528 & 0.221 & 0.357 & 7.425 & 0.169 & 0.262 & - & - & - \\
EIL~\cite{mai2020erasing} & - & 1.931 & 0.285 & 0.522 & 2.167 & 0.277 & 0.330 & - & - & - \\
TS-CAM~\cite{wsol_tr2} & DeiT & 1.842 & 0.260 & 0.336 & 2.104 & 0.201 & 0.151 & - & - & - \\
\rowcolor{gray!40}
\multicolumn{11}{c}{Weakly Supervised Affordance Grounding} \\
Hotspots~\cite{nagarajan2019grounded} & ResNet50 & 1.773 & 0.278 & 0.615  & 1.994 & 0.237 & 0.577& - & - & - \\
Cross-view-AG~\cite{AGD20K} & ResNet50 & 1.538 & 0.334 & 0.927 & 1.787 & 0.285 & 0.829 & 1.779 & 0.263 & 0.946 \\
Cross-view-AG+~\cite{luo2024grounded} & ResNet50 & 1.489 & 0.342 & 0.981 & 1.765 & 0.279 & 0.882 & 1.836 & 0.256 & 0.883 \\
LOCATE~\cite{li2023locate} & DINO & 1.226 & 0.401 & 1.177 & 1.405 & 0.372 & 1.157 & 1.593 & 0.327 & 0.966 \\
WSMA~\cite{wsma} & DINO+CLIP & 1.176 & 0.416 & 1.247 & 1.335 & 0.382 & 1.220 & 1.465 & 0.358 & 1.012 \\ 
WorldAfford~\cite{chen2024worldafford} & DINO+CLIP+SAM+GPT-4 & 1.201 & 0.406 & 1.255 & 1.393 & 0.380 & 1.225 & - & - & - \\
AffordanceLLM~\cite{qian2024affordancellm} & LLAVA-7B & - & - & - & 1.463 & 0.377 & 1.070 & - & - & - \\
\textit{Rai et al.}~\cite{rai2024strategies} & DINO+CLIP+GPT-3.5T & 1.194 & 0.400 & 1.223 & 1.407 & 0.362 & 1.170 & - & - & - \\
INTRA~\cite{jang2025intra} & DINOv2+ALBEF+GPT-4 & 1.199 & 0.407 & 1.239 & 1.365 & 0.375 & 1.209 & - & - & - \\
\hline
\rowcolor{gray!10}
\textbf{Ours} & DINO+CLIP & \textbf{1.124} & \textbf{0.433} & \textbf{1.280} & \textbf{1.243} & \textbf{0.405} & \textbf{1.368} & \textbf{1.358} & \textbf{0.378} & \textbf{1.231}  \\ \hlineB{2.5}
 \end{tabular}
 \vspace{-0.45cm}
}
\end{table*}
\endgroup

The overall process for egocentric part discovery is illustrated in Fig.~\ref{fig_pixel}.
Initially, we determine the criterion $\rho$ which is used to distinguish pixels that belong to affordable parts in egocentric images.
The logic for deriving $\rho \in \mathbb{R}^B$ is:
\begin{equation}
    \rho = \min_{e\in E} \max_{h,w\in H,W} A_{\text{obj}}^{\text{exo}}.
\end{equation}
To clarify, we first compute the maximum value across the spatial dimensions~($H\times W$) for exocentric object affinity map $A_{\text{obj}}^{\text{exo}} \in \mathbb{R}^{B\times E\times H\times W}$, resulting in a tensor of shape $B \times E$.
Then, we select the minimum value along the $E$ axis, which is the number of exocentric images paired with each egocentric image, ensuring that the weakest response among the available exocentric images is considered.
This is because exocentric images do not necessarily capture objects at a small scale; rather, some may be framed to emphasize only the specific regions involved in the interaction.
Consequently, $\rho$ is exploited to binarize the pixels of $A_{\text{obj}}^{\text{ego}}$, distinguishing affordance-relevant parts from other regions.

\noindent
\textbf{Selective Pixel Contrastive Learning.}
Part supervision within the egocentric view may not always be available in cases where exocentric images maintain a clear and unobstructed focus on target objects.
For such circumstances, object-level learning is conducted instead to distinguish target object regions against background pixels.
Thus, we utilize the hyperparameter $\gamma_2$~(equal to $\gamma_1$) to distinguish target object regions in the egocentric object affinity map $A^{\text{ego}}_{\text{obj}}$, finding that a single shared value suffices for effective separation.


Consequently, given $\mathbb{J}$ as the index set that contains indices in which the corresponding egocentric image contains pixels in object affinity map over $\rho$, positive and negative sets are organized as below:
\begin{equation}
\label{eq.selective_pixel_gathering}
\begin{split}
    Q^{+}_{b}=
    \begin{cases} 
        \{\hat{F}^{\text{ego}}_{b,h,w} | A^{\text{ego}}_{\text{obj},b,h,w} > \rho_b \} & \text{if } b \in \mathds{J}, \\ 
        \{ \hat{F}^{\text{ego}}_{b,h,w} | A^{\text{ego}}_{\text{obj},b,h,w} > \gamma_2 \} & \text{otherwise,}
    \end{cases} \\
    Q^{-}_{b}=
    \begin{cases} 
        \{\hat{F}^{\text{ego}}_{b,h,w} | A^{\text{ego}}_{\text{obj},b,h,w} \leq \rho_b \} & \text{if } b \in \mathds{J}, \\
        \{\hat{F}^{\text{ego}}_{b,h,w} | A^{\text{ego}}_{\text{obj},b,h,w} \leq \gamma_2 \} & \text{otherwise.}
    \end{cases}
\end{split}
\end{equation}
We note that the pixels in a positive set $Q^{+}_{b}$ are used as anchors for pixel contrastive learning.
Then, pixel contrastive learning is formulated as follows:
\begin{equation}
    \mathcal{L}^{\text{pix}}_{b} \!=\! \frac{-1}{|Q^{+}_{b}|^{2}}\!\!\!\sum_{z\in Q^{+}_{b}} \sum_{p \in Q^{+}_{b}} \!\log \frac{\exp(z \circ p / \tau)}{\!\!\!\!\!\!\!\!\!\!\sum\limits_{n \in (Q^{+}_{b} \cup Q^{-}_{b})} \!\!\!\!\!\!\! \exp(z \circ n / \tau)}.
\end{equation}
This encourages the model's attention to align with the discovered pixel-level clues, ensuring its attention precisely corresponds to affordance-relevant regions.


\subsection{Calibrating the Class Activation Map}
During inference, we follow previous works~\cite{li2023locate, wsma, AGD20K} to directly employ CAM as an output localization map, representing affordable regions.
However, CAM predictions often produce a Gaussian-like distribution around each salient pixel that extends beyond the actual object boundary. 
This occurs because convolution-based projection layers are utilized to encode local contexts, which spreads activations across pixels within the receptive fields.
To this end, we apply a calibration process by performing a Hadamard product between the binarized object affinity map $A$ and the CAM prediction to limit activations to only the salient parts.
Note that the process of binarization of $A$ is identical to the process of distinguishing target object regions in Eq.~\ref{eq.selective_pixel_gathering}.

\noindent\textbf{Overall Objective.}
Our objective involving classification loss, part-level prototypical contrastive loss, and pixel contrastive loss is expressed as
$\mathcal{L} = \mathcal{L}^{\text{ce}} + \lambda_1\mathcal{L}^{\text{proto}} + \lambda_2\mathcal{L}^{\text{pix}}$.


\section{Experiments}
\label{sec:experiment}

\begin{figure*}[t]
    \centering
        \vspace{-0.2cm}
    {\includegraphics[width=0.88\textwidth]{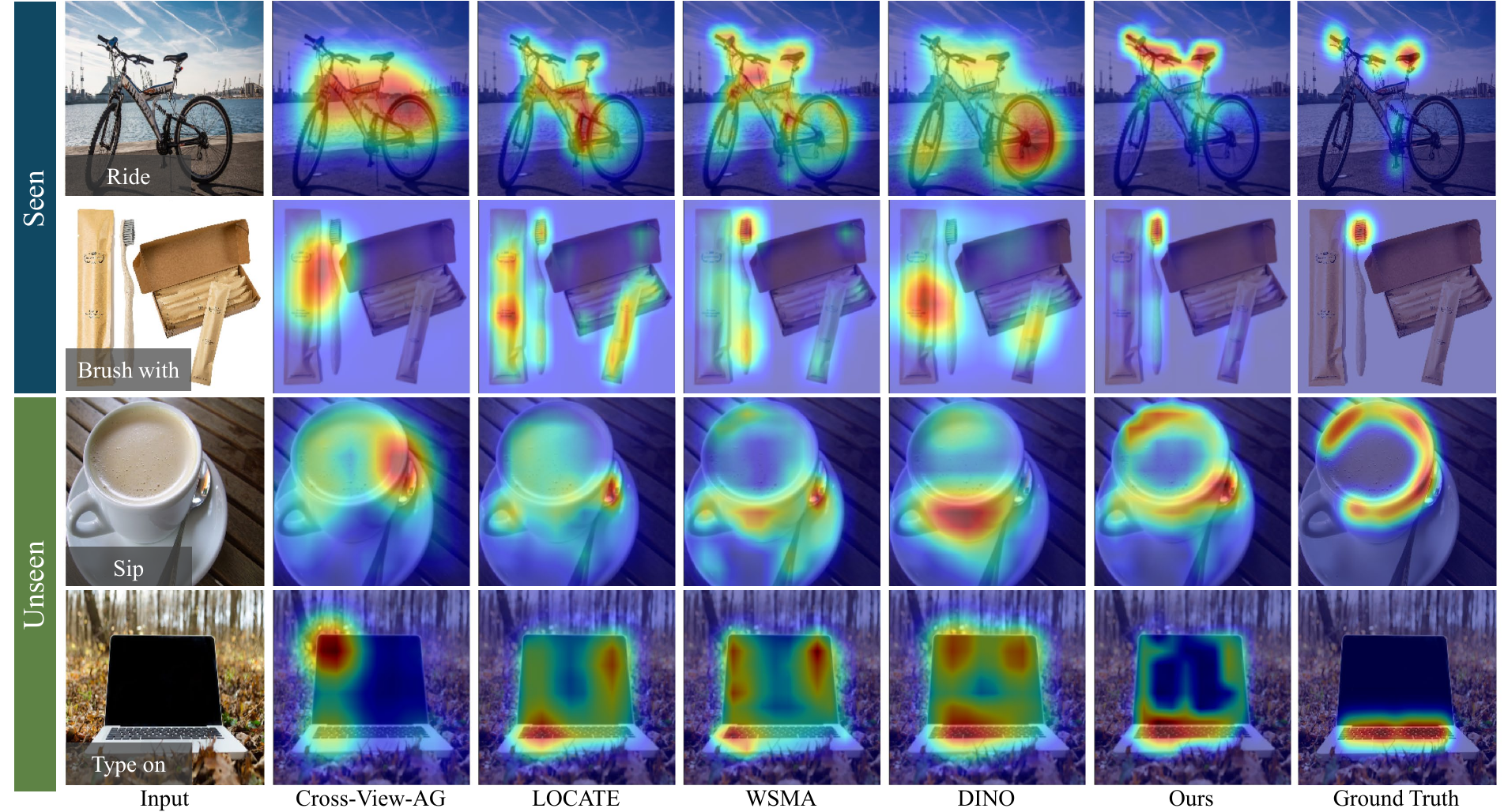} }
    \vspace{-0.2cm}
    \caption{
        Qualitative comparison results of our approach and other methods in seen and unseen domains.
    }
\label{fig_sota}
\vspace{-0.4cm}
\end{figure*}



\noindent\textbf{Evaluation Settings.}
For evaluation, we use two datasets, \textit{i.e.}, AGD20K~\cite{AGD20K}, and HICO-IIF~\cite{wsma}.
Results are evaluated with Kullback-Leibler Divergence~(KLD), Similarity~(SIM), and Normalized Scanpath Saliency~(NSS).
These metrics evaluate the similarity and the correspondence between the distributions of prediction and ground-truth heatmaps.
Also, we employ DINO ViT-S/16 and CLIP ViT-B/16 for all experiments, and set $E$~(the number of exocentric images per egocentric image) to 3, following previous works~\cite{li2023locate, wsma, chen2024worldafford, rai2024strategies}.
For hyperparameters, our loss coefficients~($\lambda_1$ and $\lambda_2$) are both set to 1.
Also, for simplicity, threshold parameters~($\alpha$ and $\gamma$) are each set to 0.6, while the bias $\beta$ and temperature $\tau$ are set to 1 and 0.5.
These hyperparameters are set the same across all datasets.
Further discussions on datasets and implementation details are in the Appendix.

\subsection{Comparison with the State-of-the-arts}
In Tab.~\ref{table_main}, we compare our proposed method against WSOL and WSAG methods utilizing various backbones~\cite{touvron2021training, CLIP, DINO,liu2024visual,li2021align,kirillov2023segment,he2016deep,achiam2023gpt}.
Models trained for object recognition, such as CLIP and WSOL methods, typically struggle with part-level grounding, as they are not optimized for identifying fine-grained affordance regions within objects. 
Yet, accurately locating affordance parts is as challenging for WSAG-tailored methods.
Thus, methods leveraging recent VLM and LLM have emerged~\cite{chen2024worldafford, qian2024affordancellm, rai2024strategies, jang2025intra}. 
Particularly, these methods often utilize LLMs to enumerate the characteristics of affordable parts within objects to improve localization with fine-grained specifications.
In this work, we follow the experimental settings of \cite{li2023locate, wsma, chen2024worldafford, rai2024strategies} to achieve a notable performance improvement across various scenarios and datasets, surpassing all previous approaches.
\begingroup
\setlength{\tabcolsep}{5.7pt} 
\renewcommand{\arraystretch}{0.9} 
\begin{table}[t]
\centering
{
 \small
 \caption{
     Study on model components.
     From left to right, we examine the benefits of object- and part-level prototypical contrastive learning~(Proto.), object- and part-level pixel contrastive learning~(Pixel.), and the calibration process with an object affinity map. 
     Cali. indicates the calibration process of the localization map. Obj. and P. denote object-level and part-level learning.
 }
 \label{table_ablation}
 \vspace{-0.1cm}
 \begin{tabular}{c|ccccc|ccc}
 \hlineB{2.5}
 \multicolumn{1}{c|}{} &\multicolumn{2}{c}{Proto.} & \multicolumn{2}{c}{Pixel.} & \multicolumn{1}{c|}{\multirow{2}{*}{Cali.}} & \multicolumn{3}{c}{AGD20K-Seen}  \\  \cline{7-9}
 \multicolumn{1}{c|}{} & Obj. & P. & Obj. & P. &  & KLD & SIM & NSS  \\ \hline
 (a) & - & - & - & - & - & 1.349 & 0.365 & 1.138 \\ 
 (b) & - & - & - & - & \checkmark & 1.271 & 0.394 & 1.162  \\ 
 (c) & \checkmark & - & - & - & - & 1.271 & 0.392 & 1.153  \\ 
 (d) & \checkmark & - & \checkmark & - & - & 1.219 & 0.402 & 1.215  \\
 (e) & \checkmark & - & \checkmark & - & \checkmark & 1.198 & 0.419 & 1.198 \\
 (f) & \checkmark & \checkmark & - & - & - & 1.164 & 0.416 & 1.290  \\
 (g) & \checkmark & \checkmark & \checkmark & - & - & 1.157 & 0.414 & 1.277  \\
 (h) & \checkmark & \checkmark & \checkmark & \checkmark & - & 1.142 & 0.415 & 1.303  \\ 
 \rowcolor{gray!10}
 (i) & \checkmark & \checkmark & \checkmark & \checkmark & \checkmark & 1.124 & 0.433 & 1.280  \\ 
 \hlineB{2.5}
 \end{tabular}
 \vspace{-0.4cm}
}
\end{table}
\endgroup

Particularly, we highlight the significant improvement in unseen scenarios, where novel objects are introduced for interaction. 
This is crucial in real-world applications, where object categories cannot be predefined.
We attribute these gains to the properties of contrastive learning.
First, our approach explicitly redirects attention away from background context and toward affordable parts/objects by enforcing contrastive objectives. 
This is especially beneficial when handling unseen objects, where the model is more prone to background distractions.
Additionally, incorporating an auxiliary self-supervised objective has been shown to enhance generalizability to novel objects~\cite{moon2022tailoring, jigen}, further strengthening the model’s robustness in diverse affordance scenarios.

Fig.~\ref{fig_sota} shows qualitative results in seen and unseen domains.
Previous works tend to identify class-wise distinguishable parts rather than focusing on affordable regions. 
For example, the bicycle frames or wheels are often highlighted instead of affordable parts for action ``ride"~(i.e., seat or handlebars).
Our approach improves affordance precision by encouraging the model to focus on affordance-relevant parts/objects while suppressing its activation on background.

\subsection{Ablation Study}
\label{sec.ablation}

\begin{figure*}[t]
    \centering
    \vspace{-0.25cm}
    {\includegraphics[width=0.92\textwidth]{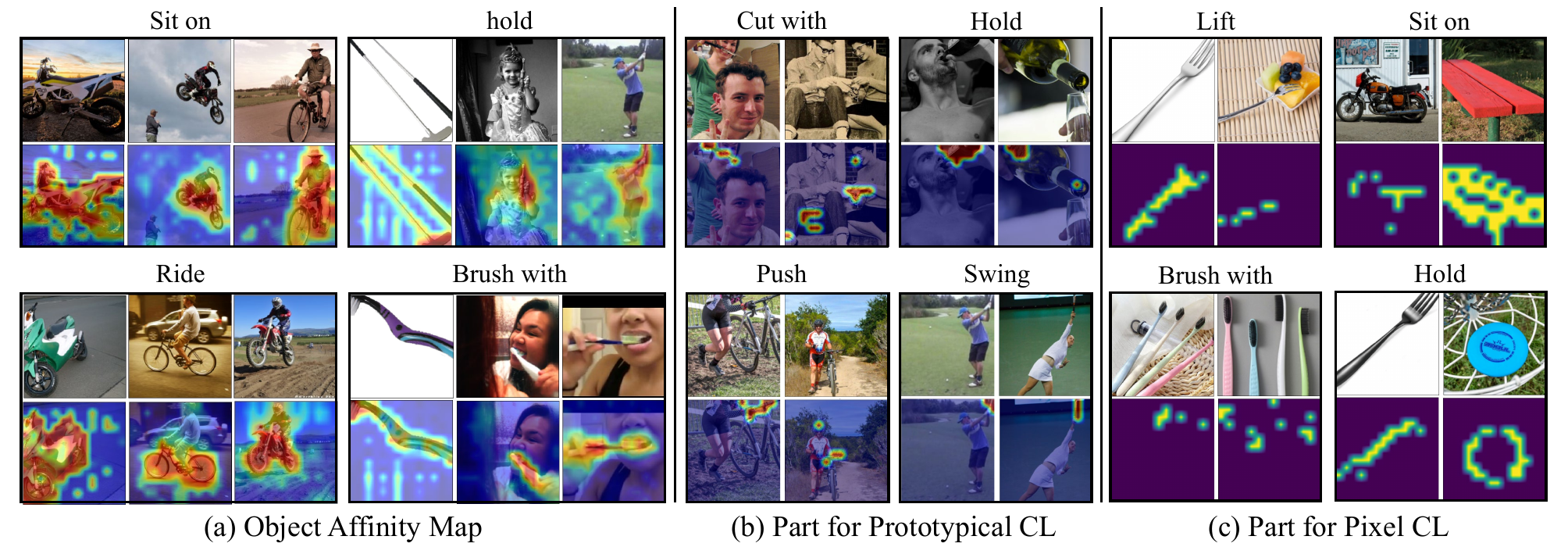} }
    \vspace{-0.2cm}
    \caption{
        Visualization of discovered objects and parts used to guide the training.
        (a) Object affinity map $A_{\text{obj}}$. 
        The leftmost sample for each class is an egocentric image and the rest are the exocentric images. 
        (b) Affordable parts of exocentric images used for prototypical contrastive learning.
        (c) Affordable parts $Q^+$ of egocentric images used for pixel contrastive learning.
    }
\label{fig_vis_pixelpart}
\vspace{-0.25cm}
\end{figure*}

Our study on component ablation is reported in Tab.~\ref{table_ablation} with fixed random seed.
For our baseline, we employ the model trained solely with the classification loss.
We then progressively integrate each learning strategy, noting that each component of our approach contributes positively to affordance grounding.
Rows (c) and (d) demonstrate the impact of object-level learning on (a), leading to significant performance improvements in cases.
These results validate our strategy to introduce object-level learning for WSAG as object-level learning mitigates the model's confirmation bias toward unaffordable but visibly distinct parts.
Additionally, the results in (f) and (h) underscore the impact of direct part-level learning, as part-level contrastive learning distinguishes affordance-relevant parts and enhances the understanding of partial details.
Finally, the calibration process with an object affinity map further improves the performances by bringing two advantages: it refines the boundary and masks out the object-irrelevant activations.


\begin{figure}
    \centering
    \vspace{-0.2cm}
    {\includegraphics[width=\linewidth]{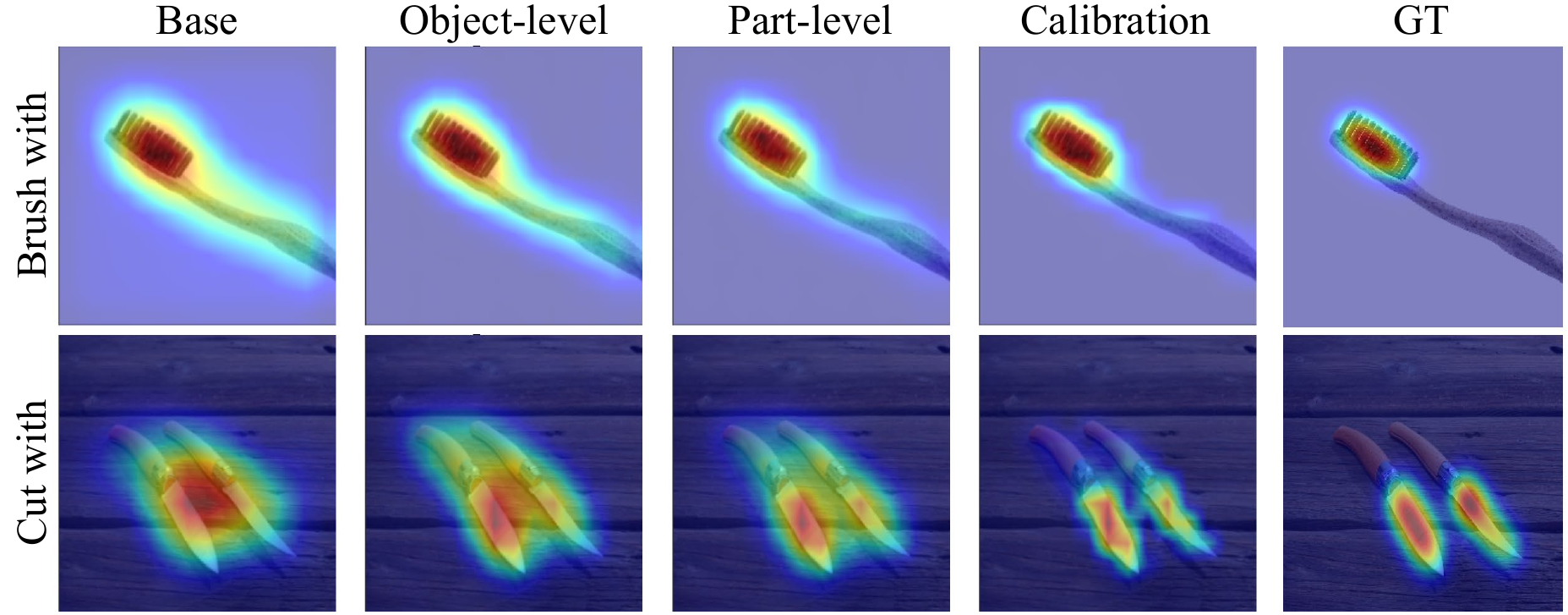} }
    \vspace{-0.4cm}
    \caption{
        Analysis of the impact of each training level, \textit{i.e.,} object and part, with qualitative results.
    }
\label{fig_ablation}
\vspace{-0.5cm}
\end{figure}
The impact of each training level is further illustrated in Fig.~\ref{fig_ablation}. 
Our baseline model tends to focus on distinct parts for each class, with activations predominantly occurring on components such as the brush and center points of knives. 
Next, we examine the effect of object-level learning, where the model spreads activations from unaffordable parts to the general object. 
When part-level learning is introduced, the activation becomes more focused on regions that are more likely to be interacted with. 
Finally, the calibration process filters out the noisy activations surrounding the salient part regions, enhancing the accuracy of affordance grounding.
These results demonstrate that the objectives of each training level are appropriately reflected.





\subsection{Study on Part and Object Level Supervision}
In Fig.~\ref{fig_vis_pixelpart}, we analyze each level of training guidance to scrutinize the benefits of our object- and part-level learning.
First, (a) displays the object affinity maps.
Despite that object affinity maps may have imprecise pixel-wise activation and only identify the foreground at a coarse granularity, we observe the accuracy in encompassing the action-associated objects.
In (b) and (c), we visualize the detected affordable parts for exocentric and egocentric images, respectively.
Specifically, (b) visualizes the detected parts within the exocentric view.
Although occasional noise is present, the identified parts generally offer reliable guidance for affordance learning.
In (c), we exhibit the identified affordable pixels $Q^+$ for part-level pixel contrastive learning on egocentric images where activated pixels generally exhibit contextual consistency.
These findings affirm that our training guidance satisfiably reflects our aim of gathering reliable supervision.

\section{Conclusion}

To enhance part-level learning, existing approaches have employed distillation strategies to guide classifiers toward affordance-relevant parts. 
Yet, since affordance cues are not always distinguishable, training is often dominated by classification, which can lead the model to focus on details frequently appearing in specific classes that may not correspond to affordable parts.
To address this issue, we introduced selective prototypical and pixel contrastive objectives that adaptively distinguish affordance-relevant cues from affordance-irrelevant regions at both the part and object levels.
Also, we introduced a part discovery algorithm to extract affordance-relevant parts within egocentric images while incorporating a modified version of an existing approach to identify parts in exocentric images. 
Lastly, we applied a localized map calibration process using an object affinity map to mitigate the activation spread caused by the receptive fields in our convolution-based CAM predictions.
Experimental results validate the effectiveness of our approach.

\section*{Acknowledgements}
\vspace{-0.2cm}
This work was supported in part by MSIT/IITP (No. RS-2022-II220680, 2020-0-01821, RS-2019-II190421, RS-2024-00459618, RS-2024-00360227, RS-2024-00437633), MSIT/NRF (No. RS-2024-00357729), KNPA/KIPoT (No. RS-2025-25393280), and SEMES-SKKU collaboration funded by SEMES.

\renewcommand{\thesection}{A}   
\renewcommand{\thetable}{A\arabic{table}}   
\renewcommand{\thefigure}{A\arabic{figure}}
\setcounter{section}{0}
\setcounter{table}{0}
\setcounter{figure}{0}

\renewcommand{\thesection}{A}   
\section{Datasets and Implementation Details}
\noindent\textbf{Datasets.} To benchmark weakly supervised affordance grounding~(WSAG) methods, we use two datasets, \textit{i.e.}, AGD20K~\cite{AGD20K}, and HICO-IIF~\cite{wsma}.
AGD20K is composed of 3,755 egocentric images with 20,061 exocentric images that belong to 36 affordance classes with 50 object classes.
Dense annotations are labeled according to the probability of interaction between the human and object regions where Gaussian blur is applied afterwards to generate the heatmaps.
HICO-IIF~\cite{wsma} comprises 1,088 egocentric images and 4,793 exocentric images.
HICO-IIF is collected from HICO-DET~\cite{chao2018learning} and IIT-AFF~\cite{nguyen2017object} where both datasets are equipped with object and affordance categories.

\noindent\textbf{Implementation Details.}
Following previous works~\cite{li2023locate, wsma}, we employ DINO ViT-S/16 for all experiments and set $E$, the number of exocentric images per egocentric image to 3.
In addition, we set $K$, the number of clusters used to segment the objects in exocentric images for part-level prototypical contrastive learning, to 3.
The model is optimized using the SGD optimizer with a learning rate of 1e-3, weight decay of 5e-4, and batch size of 8.
Additionally, while maintaining consistent parameters across datasets, we vary the number of training epochs between ADE20k and HICO-IIF. 
Specifically, we train the ADE20k dataset for 15 epochs in both seen and unseen scenarios, whereas HICO-IIF is trained for 50 epochs. 
The extended training duration~(3-4x) for HICO-IIF accounts for its dataset size, which is approximately 3–4 times smaller than ADE20k, requiring additional iterations to achieve performance saturation.
The MLP is defined with a feed-forward network and each projection layer contains two convolution layers, followed by a classifier to generate CAMs.
Projection layers for each contrastive loss are designed with a linear layer with a normalization layer.

Furthermore, as mentioned in the paper, we employ the strategy of ClearCLIP~\cite{lan2024clearclip} to enhance local discriminability in the visual features of CLIP ViT-B/16.
ClearCLIP introduces three key modifications to the original CLIP architecture in its final layer: (1) removal of the residual connection, (2) reorganization of spatial information through self-self attention~(i.e., query-to-query attention~\cite{wang2025sclip}), and (3) elimination of the feed-forward network.
These modifications are applied without the fine-tuning phase so that it uses the pretrained weights of the original CLIP.
The impact of ClearCLIP over na\"ive CLIP is shown in Tab.~\ref{table_clearclip_vs_clip}.

\renewcommand{\thesection}{B}   
\begingroup
\setlength{\tabcolsep}{9pt} 
\renewcommand{\arraystretch}{1} 
\begin{table}[t]
\centering
{
 \caption{
     Affordance grounding results using CLIP-B/16 and ClearCLIP-B/16 in the AGD20k-Seen scenario.
 }
 \label{table_clearclip_vs_clip}
 \begin{tabular}{c|c|ccc}
 \hlineB{2.5}
 Method & ZeroShot & KLD & SIM & NSS \\ \hline
 \multicolumn{1}{c|}{\multirow{2}{*}{CLIP}} & O & 1.774 & 0.250 & 0.640 \\ 
 \multicolumn{1}{c|}{} & X & 1.160 & 0.412 & 1.267 \\ \hline
 \multicolumn{1}{c|}{\multirow{2}{*}{ClearCLIP}} & O & 1.574 & 0.294 & 0.945 \\ 
 \multicolumn{1}{c|}{} & X & 1.124 & 0.433 & 1.280 \\ 
 \hlineB{2.5}
 \end{tabular}
}
\end{table}
\endgroup

\begingroup
\setlength{\tabcolsep}{1.3pt} 
\renewcommand{\arraystretch}{1} 
\begin{table}[t]
\centering
{
 \caption{
     CLIP prompt comparison in the AGD20k-Seen scenario.
    \{action\} represents the action labels.
 }
 \label{table_zeroshot}
 \begin{tabular}{c|c|ccc}
 \hlineB{2.5}
 Method & Prompt & KLD & SIM & NSS \\ \hline
 \multicolumn{1}{c|}{\multirow{2}{*}{CLIP}} & \{action\} & 1.826 & 0.242 & 0.522 \\
 \multicolumn{1}{c|}{} & ``an item to" \{action\} ``with" & 1.774 & 0.250 & 0.640 \\ \hline
 \multicolumn{1}{c|}{\multirow{2}{*}{ClearCLIP}} & \{action\} & 1.672 & 0.277 & 0.795 \\
 \multicolumn{1}{c|}{} & ``an item to" \{action\} ``with" & 1.574 & 0.294 & 0.945 \\ 
 \hlineB{2.5}
 \end{tabular}
}
\end{table}
\endgroup

\begin{figure}[t]
    \centering
    {\includegraphics[width=0.48\textwidth]{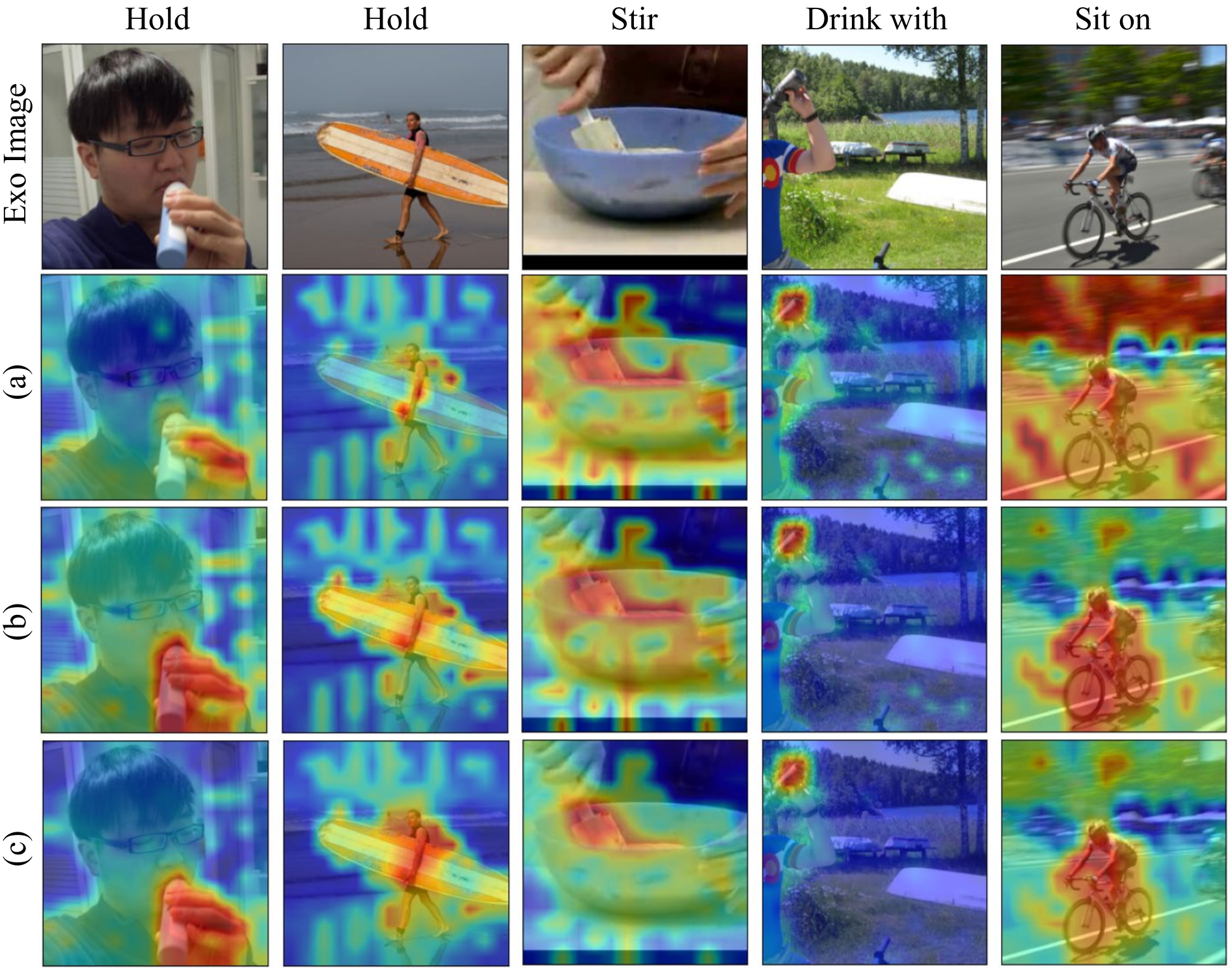} }
    \caption{
    Visualization of object affinity map for exocentric image, with various kinds of prompt.
        (a): \{action\}, (b): ``an item to'' \{action\} ``with'', (c): multiplication of ``an item to'' \{action\} ``with'' and ``a person'' \{action\} ``an item''.
    }
\label{fig_supple_exo_obj_affinity_map}
\end{figure}

\begin{figure*}[t]
    \centering
    {\includegraphics[width=0.92\textwidth]{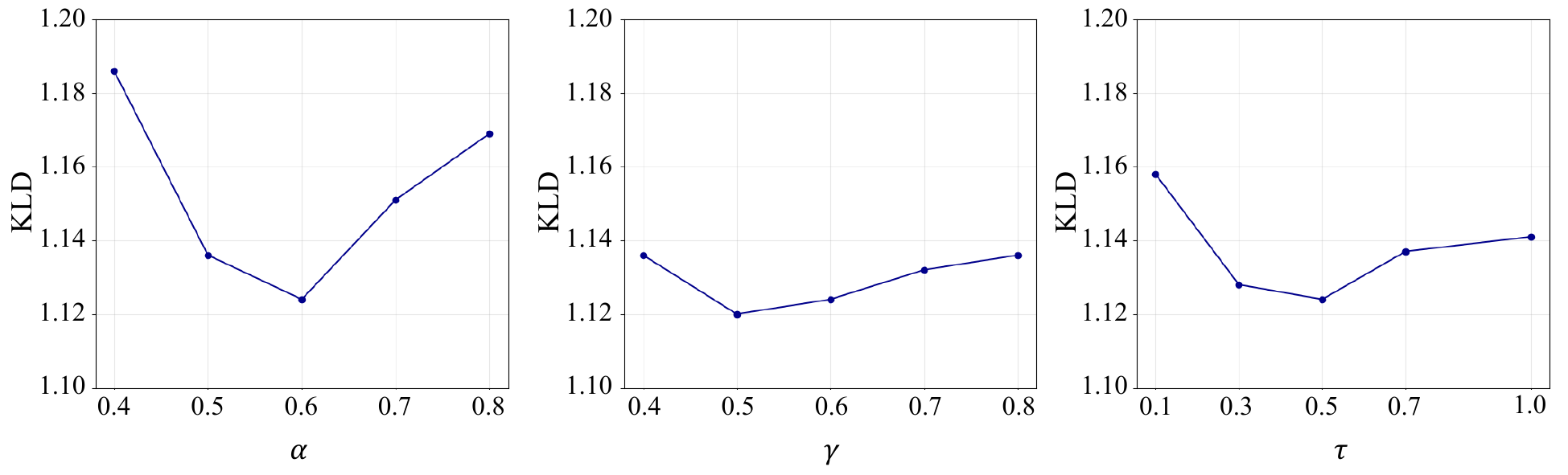} }
    \caption{
        Ablation studies of various hyperparameters. The X-axis denotes the value of each hyperparameter, the Y-axis shows the KLD performance.
    }
\label{fig_supple_hyperparam}
\end{figure*}
\section{Object Affinity Map}
In this section, we provide a detailed explanation of how the object affinity map $A$ is obtained.
Using ClearCLIP~\cite{lan2024clearclip}, we apply different strategies to infer object affinity maps for egocentric and exocentric images.

For the egocentric affinity map, we calculate the similarity between the egocentric image and action-prompted queries.
The action-prompted queries are created by augmenting the action label with a fixed prefix, ``an item to'', and a postfix, ``with''.
For example, the action label ``catch'' is augmented as ``an item to catch with''.
However, when the action label already ends with ``with'', such as ``brush with'' or ``cut with'', the postfix ``with'' is not added.
The impact of action-prompted queries is shown in Tab.~\ref{table_zeroshot}.

On the other hand, the object affinity map for exocentric images is generated using two prompting methods to focus primarily on the object parts involved in the interaction within the exocentric image, as shown in Fig.~\ref{fig_supple_exo_obj_affinity_map}.
To identify objects in exocentric images, we first use the same action-prompted queries as those applied to egocentric images, as shown in row (b) of Fig.~\ref{fig_supple_exo_obj_affinity_map}. 
However, we observe that the activation is widely distributed across the foreground objects. 
To address this, we additionally utilize entity-prompted queries to localize the entity interacting with the objects. 
We hypothesize that the intersection of the action-prompted and entity-prompted queries will yield a more accurate localization map compared to a simple similarity map derived solely from action labels.
The entity-prompted query is structured with the prefix ``a person'' and the postfix ``an item''. For example, the action label ``catch'' is augmented as ``a person catch an item''.
Yet, the similarity map obtained using the entity-prompted query may not fully capture the object parts, as the focus is on the entity in the sentence.
To address this, we apply local average pooling, which smooths the activation of each patch by averaging it with nearby patches.
Finally, we combine the affinity maps generated from the action- and entity-prompted queries by multiplying them to produce the object affinity map for exocentric images in row (c).





\renewcommand{\thesection}{C}   
\section{Hyperparameter Ablation}
We study the impact of thresholds $\alpha$ and $\gamma$ which control the reliability of selected affordable parts.
The threshold $\alpha$ determines whether the part segment within objects in exocentric images corresponds to the desired object part,
while $\gamma$ is used to binarize object affinity map of both egocentric and exocentric images into the foreground targets and the background.
Performance comparisons for varying $\alpha$ and $\gamma$ are illustrated in Fig.~\ref{fig_supple_hyperparam}.
Our results indicate that $\alpha$, used for selecting reliable clusters~(groups of pixels), is more sensitive than $\gamma$.
However, both thresholds consistently achieve optimal performance within the range of 0.5 to 0.6.
In this work, we set $\alpha$ and $\gamma$ to 0.6.

Additionally, we examine the effects of varying $\tau$, the scaling parameter used in both prototypical and pixel contrastive losses.
Results are shown on the right side of Fig.~\ref{fig_supple_hyperparam}.
In this work, we set $\tau$ to 0.5 as it outcomes the best result.

Although the performance slightly decreases when adjusting our hyperparameters, our results demonstrate the robustness of the framework.
In particular, our model consistently achieves state-of-the-result performances regardless of hyperparameters $\alpha$, $\gamma$, and $\tau$.

\begin{figure}[t]
    \centering
    {\includegraphics[width=0.47\textwidth]{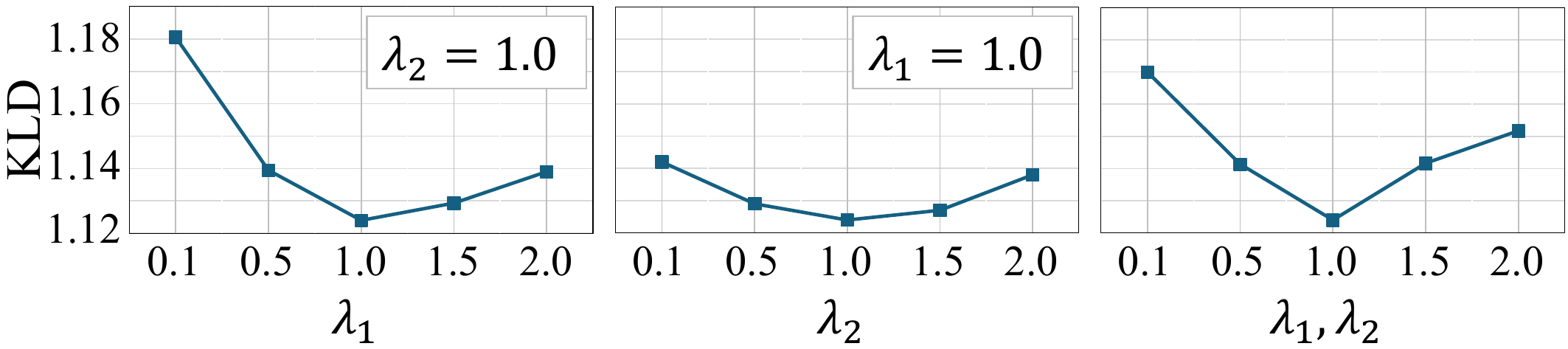} }
    \caption{
    Study on loss coefficients. $\lambda_1$ and $\lambda_2$ are coefficients for prototypical and pixel contrastive learning, respectively.
    We vary each coefficient while keeping the others fixed at their default value of 1 and also examine their impact when adjusted simultaneously.
    }
    \label{fig_lossablation}
\end{figure}

Study on loss coefficients are in Fig.~\ref{fig_lossablation}.
As shown, our default value of 1 yields its best result.
Nevertheless, our proposed approach consistently outperforms baselines by a significant margin, demonstrating its robustness and insensitivity to extensive parameter tuning.

\renewcommand{\thesection}{D}   
\begin{figure*}[t]
    \centering
    {\includegraphics[width=1.\textwidth]{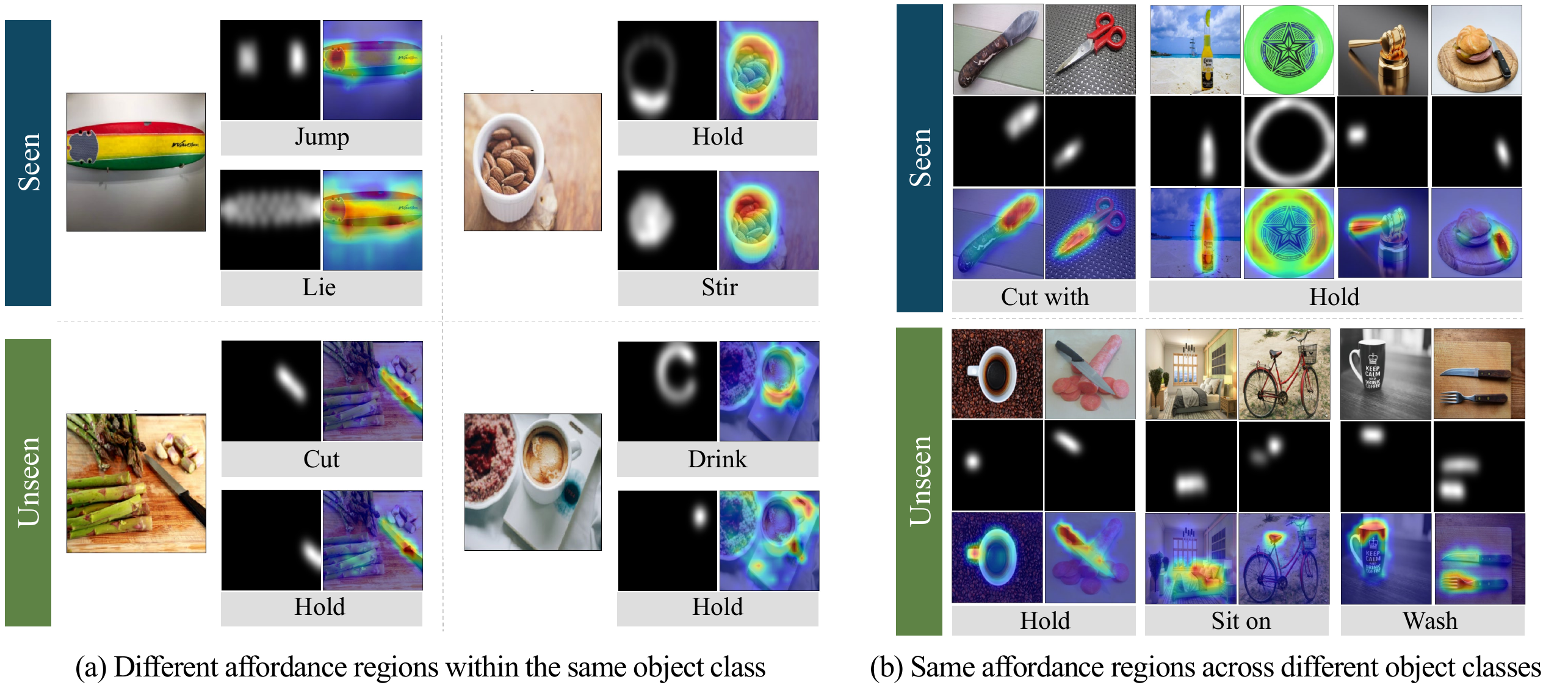} }
    \caption{
        Visualization of the test image, ground-truth label, and our prediction on AGD20K dataset.
    }
\label{fig_supple_objectbias_figure}
\end{figure*}
\section{Bias on Object and Affordance Classes}
Objects can be involved in various actions, and likewise, different affordance classes may occur across diverse objects. 
This presents a particular challenge in weakly supervised affordance grounding, where the distinctions between classes are not explicitly provided. 
In Fig.~\ref{fig_supple_objectbias_figure}, we examine how our proposed approach performs under such scenarios.
First, Fig.~\ref{fig_supple_objectbias_figure}~(a) illustrates the prediction results when different affordance classes are queried for the same object class. 
While the predictions are not perfectly accurate, the model still exhibits meaningful distinctions between affordance classes despite the absence of explicit class-level cues.
Fig.~\ref{fig_supple_objectbias_figure}~(b) further visualizes how well the model generalizes affordance understanding across diverse object classes, demonstrating notably consistent performance.
These results support that our strategy effectively minimizes biases toward specific object–affordance pairings, promoting robust affordance predictions.



\renewcommand{\thesection}{E}   
\begingroup
\setlength{\tabcolsep}{4pt} 
\renewcommand{\arraystretch}{1.0} 
\begin{table}[t]
\centering
{
 \caption{
     Comparison results between DINO attention map and CLIP affinity map to measure pIoU.
 }
 \label{table_nodinoattn}
 \begin{tabular}{c|c|ccc}
 \hlineB{2.5}
 Dataset-Scenario & Method & KLD$\downarrow$ & SIM$\uparrow$ & NSS$\uparrow$ \\ \hline
 \multirow{2}{*}{AGD20K-Seen} & DINO-attn & 1.124 & 0.433 & 1.280 \\ \cline{2-5}
 & CLIP-obj. & 1.126 & 0.435 & 1.273 \\ \hline
 \multirow{2}{*}{AGD20K-Unseen} & DINO-attn & 1.243 & 0.405 & 1.368 \\ \cline{2-5}
 & CLIP-obj. & 1.257 & 0.398 & 1.360 \\
 \hlineB{2.5}
 \end{tabular}
}
\end{table}
\endgroup

\section{DINO Attention Map for Prototype Selection}

In prototype generation for prototypical contrastive learning, we utilize the self-attention map from DINO to measure pIoU, which allows us to select the most suitable prototype among three candidates and perform part-level learning. We emphasize that the DINO attention map can be replaced by any alternative capable of identifying the main object within egocentric images. To validate this flexibility, we conduct experiments using the CLIP affinity map as an alternative, applying a specific threshold~(0.75) to distinguish foreground from background regions. Table~\ref{table_nodinoattn} compares the results obtained using DINO attention maps and CLIP affinity maps, demonstrating the robustness and versatility of our method.

\renewcommand{\thesection}{F}

\begin{figure*}[t]
    \centering
    {\includegraphics[width=0.85\textwidth]{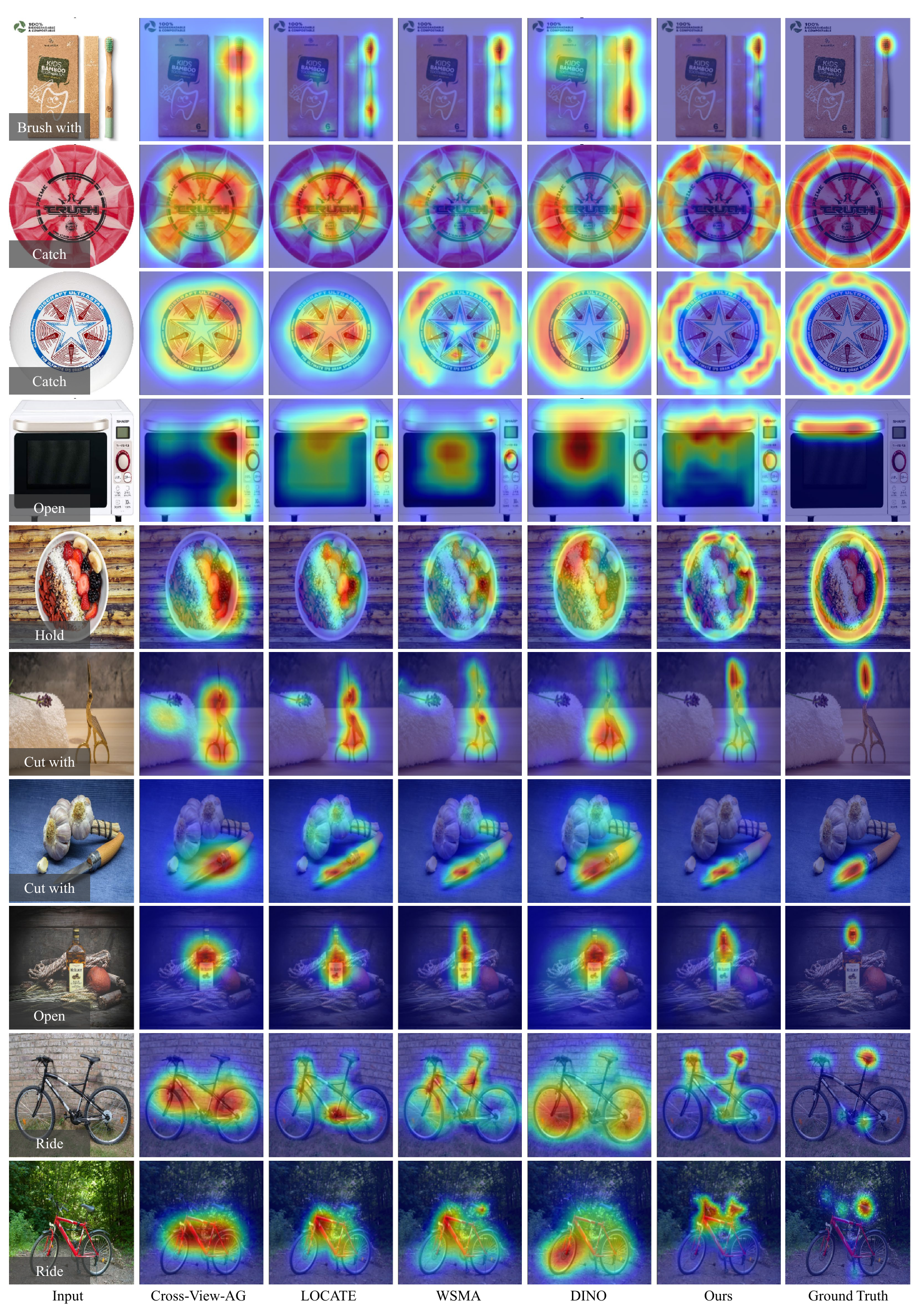} }
    \caption{
        Affordance grounding results of our approach and other methods in the seen domain.
    }
\label{fig_supple_qual}
\end{figure*}
\begin{figure*}[t]
    \centering
    {\includegraphics[width=0.85\textwidth]{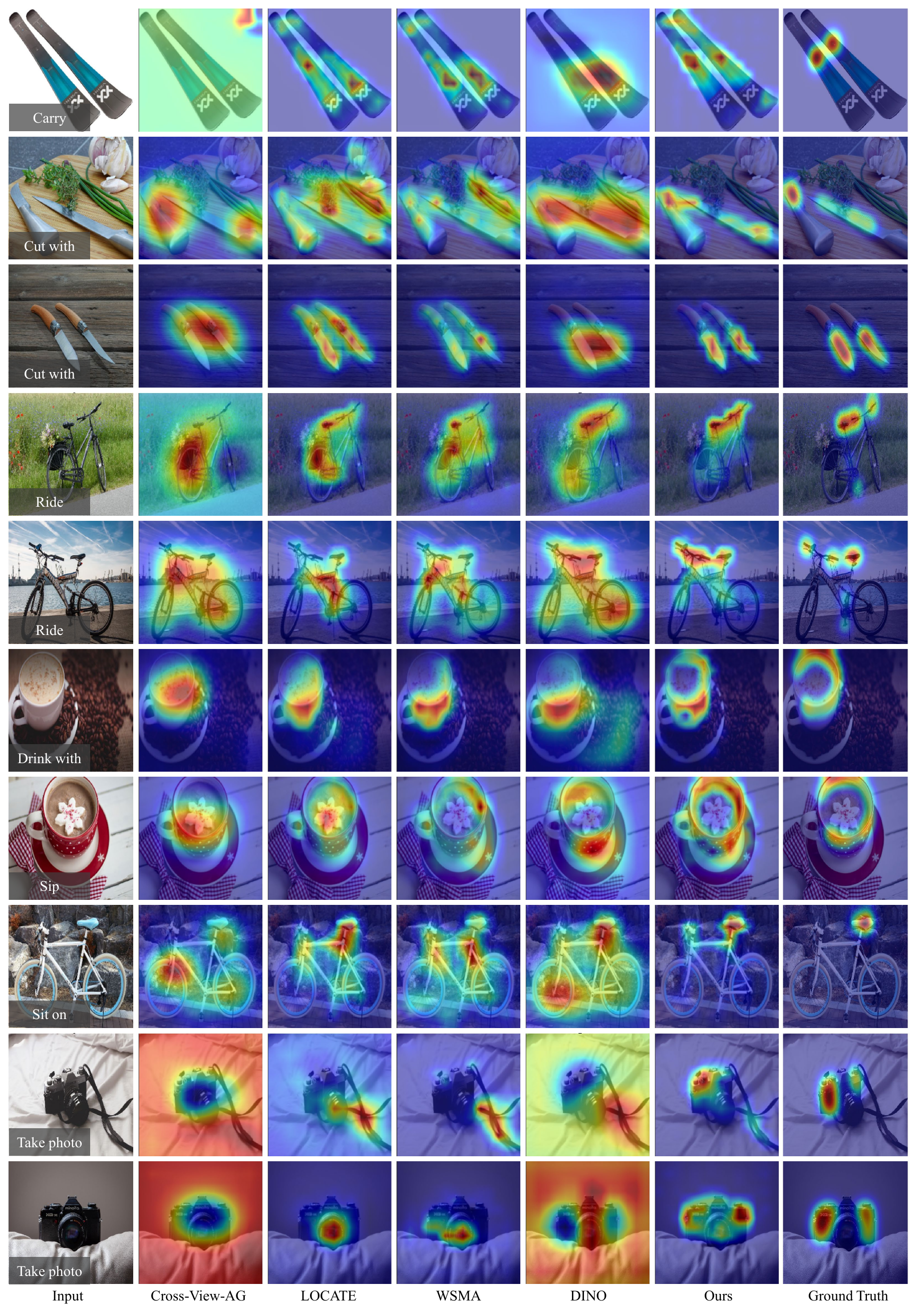} }
    \caption{
        Affordance grounding results of our approach and other methods in the unseen domain.
    }
\label{fig_supple_qual_unseen}
\end{figure*}

\section{Additional Qualitative Results}
Additional qualitative results in comparison to baseline methods are depicted in Fig.~\ref{fig_supple_qual} and Fig.~\ref{fig_supple_qual_unseen}.
Particularly, Fig.~\ref{fig_supple_qual} illustrates the results in the seen domain, while Fig.~\ref{fig_supple_qual_unseen} focuses on the unseen domain.
As observed, we find that our proposed approach consistently demonstrates more accurate results than previous works.

\newpage

{
    \small
    \bibliographystyle{ieeenat_fullname}
    \bibliography{main}
}

\end{document}